# Hardware Acceleration for Boolean Satisfiability Solver by Applying Belief Propagation Algorithm

Te-Hsuan Chen and Ju-Yi Lu
Department of Electrical Engineering and Computer Science
University of Michigan, Ann Arbor, MI, 48109, USA
{tehsuan, juyi}@umich.edu

## ABSTRACT
Boolean satisfiability (SAT) has an extensive application domain in computer science, especially in electronic design automation applications. Circuit synthesis, optimization, and verification problems can be solved by transforming original problems to SAT problems. However, the SAT problem is known as an NP-complete problem, which means there is no efficient method to solve SAT problems. Therefore, the design of an efficient SAT solver to enhance the performance is always desired. In this paper, we proposed a hardware acceleration method for Boolean SAT problems. By surveying the properties of SAT problems and the decoding of low-density parity-check (LDPC) codes, a special class of error-correcting codes (ECCs), we discover that both of them are constraint satisfaction problems (CSPs). The belief propagation algorithm (BPA) has been successfully applied to the decoding of LDPC, and the corresponding decoder hardware designs are extensively studied. Therefore, we proposed a belief propagation based algorithm to solve SAT problems. With this algorithm, the SAT solver can be accelerated by hardware. A software simulator is implemented to verify the proposed algorithm and the performance improvement is estimated. Our experiment results show that time complexity does not increase with the size of SAT problems and the proposed method can achieve at least 30× speedup compared to MiniSat.

## Keywords
SAT, Boolean Satisfiability, Belief Propagation Algorithm.

## 1. INTRODUCTION
Boolean satisfiability (SAT) is a class of problems that establishes if there exist an assignment to variables of a Boolean formula that evaluates it to true [1]. This Boolean formula is usually in the conjunctive normal form (CNF), which is an expression of a conjunction (AND, ∧) of clauses. Each clause is a disjunction (OR, ∨) of literals, and each literal is either a variable or its negation (NOT, ¬). SAT has an extensive application domain. In logic design, synthesis, optimization and verification, many problems, such as placement and route, test pattern generation, and equivalence checking, and model checking, etc., can be expressed as variables whose values are in the set $\{true, false\}$ or $\{1,0\}$.

The SAT problem is the first known NP-complete decision problem [2], which means that there is no efficient algorithm to solve SAT problems. Although theoretically, finding the set of assignment can be achieved by exhaustively trying all the possible assignments, this method becomes infeasible quickly as the number of literals and clauses increases. The set of algorithms that can solve SAT problems are called SAT solvers. However, due to the NP-complete nature of SAT, how to solve SAT problems within a reasonable time becomes an important issue. Much research has been conducted to improve the efficiency of SAT solvers, but no algorithm so far can efficiently solve all SAT instances.

The goal of this paper is using hardware to accelerate SAT solvers. The approaches to solve the SAT problem have divided into software solvers, which solving the problem using computer simulation, and hardware solver, to implement Verilog code running on FPGA or virtual machine [7][10][11]. For hardware SAT solver, the main two categories are instance-specific (instance dependence) and application-specific. In this work, we try to implement an instance-specific model, which is more resource efficient to accommodate on a single FPGA.

To satisfy a CNF, i.e. to find a set of value that makes the CNF=1, each clause must be 1 too. Therefore, each clause represents a constraint that must be satisfied simultaneously. A constraint satisfaction problem (CSP) consists of three components, $X$, $D$, and $C$ [3]:

1. $X$ is a set of variables, $\{X_1, X_2, \ldots, X_n\}$.
2. $D$ is a set of domains, $\{D_1, D_2, \ldots, D_n\}$, one for each variable.
3. $C$ is a set of constraints that specify allowable combinations of values.

Each domain $D_i$ consists of a set of allowable values, $\{v_1, v_2, \ldots, v_k\}$ for variable $X_i$. Each constrain $C_i$ consists of a pair $\langle scope, rel \rangle$, where $scope$ is a tuple of variables that participate in the constrain and $rel$ is the relation that defines the values that those variables can take on. A problem is solved when each variable has a value that satisfies all constraints on the variable.

From the definition of the CSP, it can be found that Boolean SAT is a special case of the CSP, where the set of domains is $\{true, false\}$ or $\{1,0\}$ and the constraints are defined by a set of clauses that must be satisfied simultaneously.

Error correction codes (ECC's) provide one of the most cost-effective ways to achieve noise protection. By applying the ECC to data, errors can be corrected or detected. An (n, k) code maps a binary k-tuple called a message block to a binary n-tuple called a codeword block. Usually, message and codeword blocks are represented in vector forms so the mapping is defined by a set of matrix multiplications. The decoding of a special class of ECC, low-density parity-check (LDPC) codes [4], is a constraint satisfaction problem. The parity-check equations are specially designed equations that are equal to zero if all the inputs are correct.

The goal of decoding LDPC codes is to find a set of input assignments such that all the parity-check equations are equal to 0. Therefore, the decoding of LDPC codes is actually a CSP. Each symbol of the codeword is a variable; each parity check corresponds to a constraint in CSP. Besides, the domain of the variables is limited to $\{true, false\}$ or $\{1,0\}$ in LDPC codes. Therefore, the decoding of LDPC codes can be modeled as a SAT problem.

In the decoding of LDPC codes, the procedure can be illustrated by a bipartite graph which contains two types of nodes: variable nodes and parity-check nodes, and use iterative belief propagation decoding method to find solutions [8][9]. Since both the decoding of LDPC codes and SAT solvers are CSPs, and the decoding of LDPC codes can be accomplished by specific designed hardware. Here comes the question: can we extend the LDPC algorithm, belief propagation algorithm (BPA), to solve SAT problems so that a SAT solver can be speeded up by hardware? In this paper, we show that BPA can also be applied to SAT solvers since both of them solve CSPs. It is also possible a hardware circuit using similar architecture can also be used to accelerate SAT solvers. Therefore we propose a method to utilize a hardware based on LDPC decoder to improve the speed to SAT solvers.

## 1.1 Contribution

Our main contribution in this paper is (1) applying the algorithm of decoding LDPC codes, to CNF-based SAT problems, (2) evaluating the proposed modified BPA by a simulator written in C/C++, (3) estimating the time that is need by the proposed SAT solver and comparing it with MiniSat [5].

## 1.2 Organization

The paper is organized as follows. Section 2 gives a quick review on the LDPC codes and the belief propagation algorithm. A SAT solving algorithm based on the belief propagation algorithm is proposed in section 3. Section 4 provides the experimental results. Section 5 discusses the related work and Section 6 gives the conclusion.

## 2. LOW-DENSITY PARITY CHECK CODE

Originally proposed in 1962 by Robert Gallager [4] and re-discovered by David MacKay in 1996 [6], LDPC codes with iterative decoding algorithms have performance closely approaching Shannon channel capacity. LDPC codes went unnoticed until the 1990s because the hardware could not support effective decoder implementations. However with MacKay's research and the development of semiconductor technology, the hardware implementation of LDPC codes became feasible.

A LDPC code is a linear block code for which the parity check matrix of interest has a low density of ones. A LDPC code can be represented by matrices. Assume that $C$ is the codeword set of a *(n, k)* binary linear block code. If a $k \times n$ matrix $G$ is its generation matrix, then $C = \{\vec{x} \cdot G | \vec{x} \epsilon \{0,1\}^k\}$ for any length-$k$ message $\vec{x}$. An $m \times n$ matrix $H$ is its parity check matrix, if $C = \{\vec{c} | \vec{c} \epsilon \{0,1\}^n, \vec{c} \cdot H^T = 0 \}$ where $\vec{c}$ is a codeword of $C$. In other words, an LDPC code is designed to satisfy that for every codeword $\vec{c} \in C$, this equation $\vec{c} \cdot H^T = 0$ always holds.

LDPC codes can also be represented by using graphs. A parity-check matrix $H$ can be associated with a graph called bipartite graph. This graph contains two set of nodes: variable nodes and check nodes (*v*-nodes and *c*-nodes). The rule for constructing the graph is that a *c*-node $i$ is connected to a *v*-node $j$ whenever element $h_{ij}$ in $H$ is a 1. For example, let the parity-check matrix $H$ be a 3×4 matrix

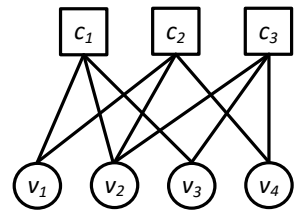

The corresponding bipartite graph is shown in Figure 1. There are 4 *v*-nodes and 3 *c*-nodes in this example.

**Figure 1: Example of bipartite graph.**

Besides, a cycle of length $l$ in a bipartite graph is a path comprised of $l$ edges from a node back to the same node. For example the bipartite Graph of Fig. 1 has a cycle of length 6. Short cycles are usually considered bad in graphs used for iterative decoding based on "*message passing*;" they increase the dependence of information being received at each node during message passing which may lead to latch-up problems. LDPC codes depend on a message passing decoding algorithm called belief propagation algorithm (BPA) or sum-product algorithm (SPA). The BPA is an efficient decoding algorithm which iteratively decodes LDPC code based on belief propagation. BPA decodes data with the probability computation of received signals based on the characteristic of the channel. The input bit probabilities are called the a priori probabilities for the received bits because they were known in advance before running the LDPC decoder. Besides, two sets of probabilities computation are handled in the decoder. One is related to the decoding criterion which determines the value of a received bit from the value of a received signal. The sets of probabilities are referred to as posterior probabilities.

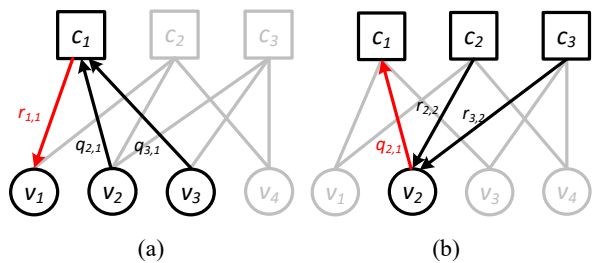

**Figure 2: Iterative update of the BPA: (a) from a check node to a variable node, (b) form a variable node to a check node.**

The other probability is related to the satisfaction of each check sum given the received signal on each bit. BPA decoding can decode the data by iterative update in these two sets of probabilities until the satisfaction of the received bits is determined by the first sets of probabilities in the parity-check equations. The goal of the decoder is to find the maximum a posteriori probability (MAP) for each codeword bit. In other words, the aim of SPA is to maximize the probability that all parity-check constraints are satisfied for each codeword bit.

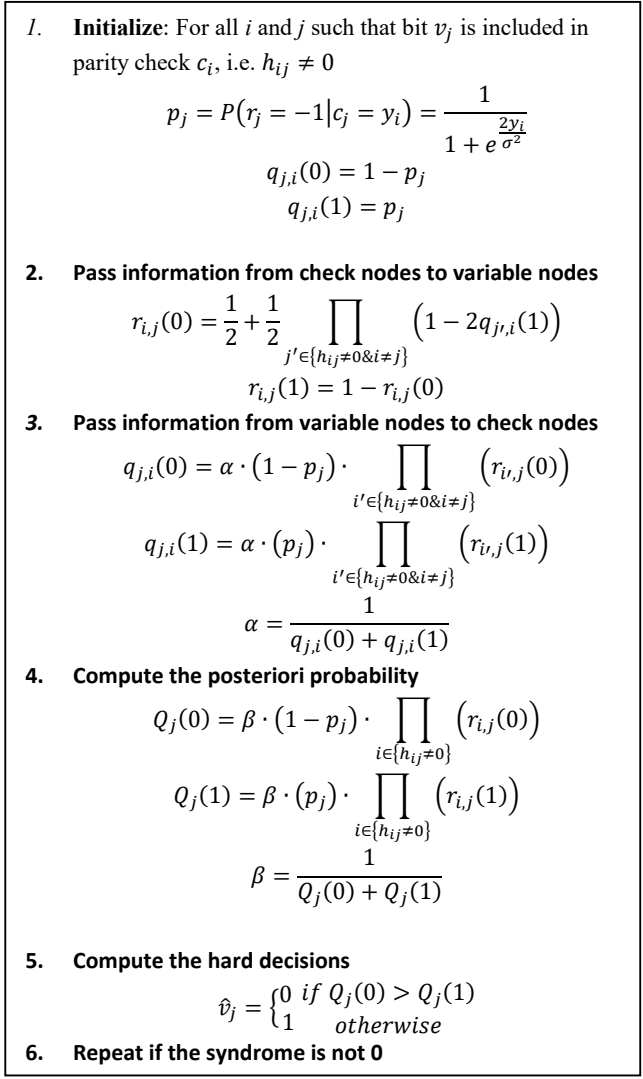

1. **Initialize**: For all $i$ and $j$ such that bit $v_j$ is included in parity check $c_i$, i.e. $h_{ij} \neq 0$

$$p_j = P(r_j = -1 | c_j = y_i) = \frac{1}{1 + e^{\frac{2y_i}{\sigma^2}}}$$
$$q_{j,i}(0) = 1 - p_j$$
$$q_{j,i}(1) = p_j$$

2. **Pass information from check nodes to variable nodes**
$$r_{i,j}(0) = \frac{1}{2} + \frac{1}{2} \prod_{j' \in \{h_{ij} \neq 0 \& i \neq j\}} (1 - 2q_{j',i}(1))$$
$$r_{i,j}(1) = 1 - r_{i,j}(0)$$

3. **Pass information from variable nodes to check nodes**
$$q_{j,i}(0) = \alpha \cdot (1 - p_j) \cdot \prod_{i' \in \{h_{ij} \neq 0 \& i \neq j\}} (r_{i',j}(0))$$
$$q_{j,i}(1) = \alpha \cdot (p_j) \cdot \prod_{i' \in \{h_{ij} \neq 0 \& i \neq j\}} (r_{i',j}(1))$$
$$\alpha = \frac{1}{q_{j,i}(0) + q_{j,i}(1)}$$

4. **Compute the posteriori probability**
$$Q_j(0) = \beta \cdot (1 - p_j) \cdot \prod_{i \in \{h_{ij} \neq 0\}} (r_{i,j}(0))$$
$$Q_j(1) = \beta \cdot (p_j) \cdot \prod_{i \in \{h_{ij} \neq 0\}} (r_{i,j}(1))$$
$$\beta = \frac{1}{Q_j(0) + Q_j(1)}$$

5. **Compute the hard decisions**
$$\hat{v}_j = \begin{cases} 0 & \text{if } Q_j(0) > Q_j(1) \\ 1 & \text{otherwise} \end{cases}$$

6. **Repeat if the syndrome is not 0**

**Figure 3: BPA for decoding ECC.**

Figure 2 illustrate the basic concept of the BPA. Figure 2(a) is a part of a bipartite graph; it only shows nodes and edges that connected to a parity-check $c_1$. Here the relation between $c_1$ and $v_1$, $v_2$, and $v_3$ is: $c_1 = v_1 \oplus v_2 \oplus v_3$ where $\oplus$ is XOR. To satisfy this parity-check, $v_1$, $v_2$, and $v_3$ must be all 0's or only two of them are 1's. The BPA handles two sets of probabilities. The first set is related to the decoding criterion. These quantities are defined and iterative computed in the BPA:

(1) $r_{i,j}(k)$: The probability $c_i = 1$ given $v_j = k$ and probabilities of other variables, $q_{ij}$'s.

(2) $q_{j,i}(k)$: The probability $v_j = k$ when all parity checks involving $v_j$ except $c_i$ are satisfied.

For example, $r_{1,1}(1)$ is the probability of $c_1$ =1 given the information $q_{2,1}$ and $q_{2,1}$. Note that since $c_1 = v_1 \oplus v_2 \oplus v_3$ if $q_{2,1}$ and $q_{2,1}$ are high, i.e. the probability that $v_2$ and $v_3$ are 1's are high, the probability that $v_1 = 0$ must be high to satisfy $c_1 = 0$.

On the other hand, for the $v$-node, Figure 2(b) is a partial graph that shows all the parity-check nodes that use $v_2$. The $v$-node $v_2$ receives the information of probabilities that satisfy all the parity-checks that $v_2$ involves in and then uses this information to update its probability. For example, in Figure 2(b) $v_2$ passes its probability to $c_1$ by using the information form $c_1$ and $c_3$.

Another set of probability that is related to the decoding criterion is also calculated. It is denoted as $Q_j$, representing the probability that all checks involving $j^{\text{th}}$ bit are satisfied. This probability is also called pseudo-posterior probability.

There are five steps in the BPA shown in Figure 3 [12]. The original algorithm is designed for a communication system assuming signals are modulated in binary phase-shift keying (BPSK) where logic 1 is modulated as -1 voltage and logic 0 is 1 voltage. The noise is assumed to be additive white Gaussian noise (AWGN). The general form of the BPA is as follows:

## 3. BELIEF PROPAGATION FOR SAT

This section we discuss the proposed Belief Propagation Algorithm for SAT problem (BPA-SAT) for SAT problems. In 3.1 we discuss the adopted modification on the Belief Propagation to apply on the SAT solver. In 3.2 we provide an overview of proposed BPA-SAT and some detail implementation.

### 3.1 Adaption on BPA-SAT

BPA can be adopted on SAT problem once we treat the check nodes as the clauses, and variable nodes as literal. Also, the initialization of the probability is less important than the conventional BPA, because there is no prior information about the literal assignment.

1. **Initialization**: in the SAT problem, the initial values of literals are all set to 0.5 to show no bias of the assignment without any prior information.

2. **Pass information from clauses to literals**: Since the operators in the CNF clauses are OR, which is $c_1 = v_1 + v_2 + v_3$, $r_{1,1}(1)$ represents the probability of $c_1$ is true given the $v_1 = 0$, and the probability of $v_2$, $v_3$. Therefore the modified equation is:

$$r_{1,1}(0) = P(c_1 = 1 | v_1 = 0, \vec{q})$$

$\vec{q}$ represents the vector of $q_{i,j}(1)$.

3. **Pass information from literals to clauses:** We neglect the multiplication on $(1 - p_j)$ or $p_j$, because the initial values should not affect the value since they don't contain useful information.

$$q_{j,i}(1) = \alpha \prod_{i' \in \{h_{ij} \neq 0 \& i \neq j\}} (r_{i',j}(1))$$

4. **Compute the posteriori probability:** Same as above, remove the multiplication on $(1 - p_j)$ or $p_j$.

$$Q_j(1) = \beta \prod_{i \in \{h_{ij} \neq 0\}} (r_{i,j}(1))$$

## 3.2 BPA-SAT overview

Figure 4 is the flow chart of the whole BPA-SAT. There are two loop in BPA-SAT, the inner loop is the core algorithm of BPA-SAT, which iteratively calculates $r_{i,j}(1)$, $q_{i,j}(1)$, and $Q_j(1)$. The outer loop is the break-and-restart mechanism to prevent BPA-SAT running forever, and randomly assign another start point (initial value set).

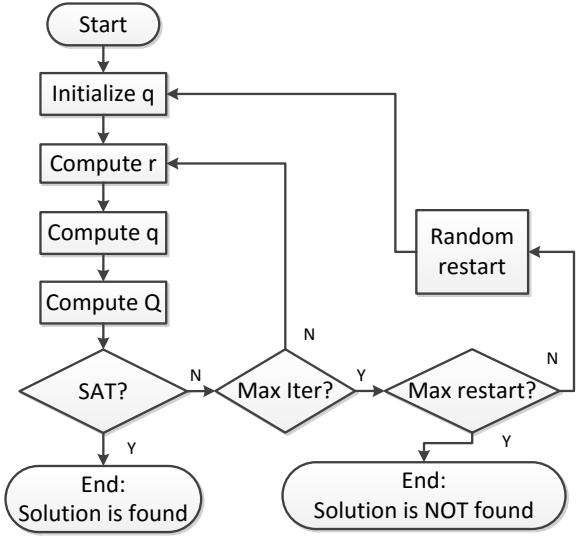

**Figure 4: Flow chart of the BPA for CNF-based SAT problems**

## 4. EXPERIMENTAL RESULT

In this section we present an evaluation of completeness and speedup of our belief propagation for SAT problem. The 3SAT benchmarks in SATLIB [12] are ran on our BPA for SAT simulator. For the speedup, we compares the estimated execution time of BPA for SAT running on FPGA Xilinx Virtex 2 XC2V6000-5 and the real execution time of MiniSat solver running on two 6-core Intel(R) Xeon(R) CPU E5645 @ 2.40GHz processor.

### 4.1 Completeness

In this sub-section we investigate the completeness of the BPA-SAT. In our experiment, all the instances in the benchmark are satisfiable, completeness is defined as the fraction of testbench instance that can be solved by the BPA-SAT.

Figure 5 examines the completeness of BPA for SAT. All the benchmark set are satisifiable. Benchmark uf20-91 indicates the literal number 20 and clause number 91. It shows that BPA for SAT has a worse solving ability when the problem size become larger. After adopting the random restart mechanism, there is only an average 5% improvement. Therefore, how to improve the completeness will be the main issue in the future works.

Moreover, the ECC decoder where BPA works well has the literal-clause ratio greater than 1, while the literal-clause ratio of SAT problem are typically smaller than 0.25. This fundamental difference is believed to be a crucial factor of the incompleteness in the proposed BPA for SAT.

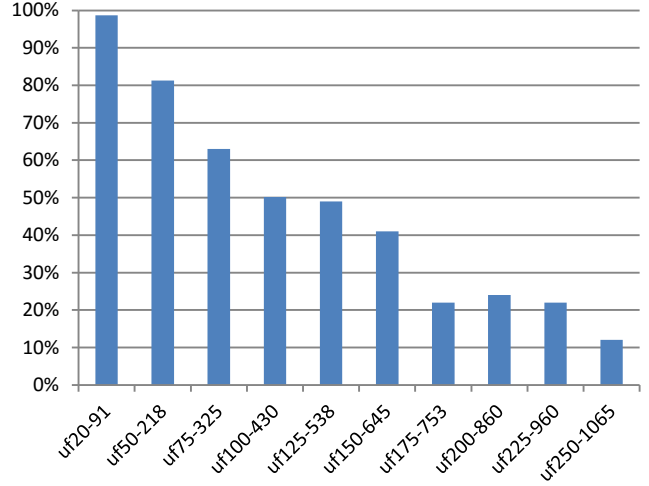

**Figure 5: The completeness of BPA-SAT.**

### 4.2 Execution iteration & Speedup Evaluation

To estimate the speedup of the algorithm, we take the FPGA Xilinx Virtex 2 XC2V6000-5 as reference model. Based the experimental result of [13], the desired FPGA computation ability, second per iteration (SPI) can be calculated in equation below:

$$SPI = \frac{Codeword\ length}{Throughput \times Iteration} = \frac{3969}{1417M \times 15}$$
$$= 1.86 \times 10^{-7} sec/iteration$$

In [13], the *Codeword length* is the bandwidth the FPGA can execute in parallel in a single iteration. The *Iteration* is the experimental result of the iteration number.

Therefore the estimated speedup can be derived as below:

$$\textbf{Speedup} = T_{minisat}/T_{exec} = T_{minisat}/(iteration \times SPI) \quad (1)$$

$T_{minisat}$ is the average running time for the MiniSat solver to solve a single SAT instance. $T_{exec}$ is the estimated execution time of BPA for SAT to running on FPGA.

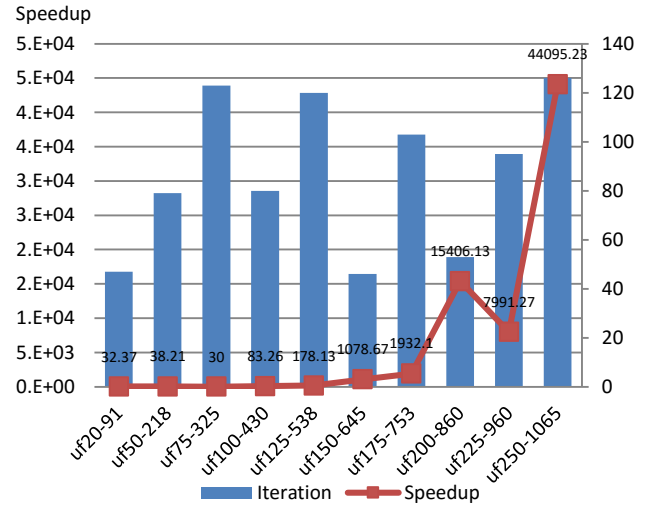

**Figure 6: Average iteration and the estimated speedup.**

Figure 6 takes a closer look on every solved instance to find the relation between propagation iteration and the scale of the SAT problem. The blue histogram indicates the average iteration of the solved instances. The line chart represents the average speedup compared with the execution time of MiniSat SAT solver. The average iteration number of all the solved instance doesn't show any trend of growth as the growth of the problem size, which imply the constant timing complexity regardless of the scale of the SAT problem.

The line chart comes from the equation (1). The speedup grows in exponential, because the execution time of MiniSat grows in exponential. The spike on uf200-860 is caused by the large standard deviation when the sample number is small. This chart indicates the main advantage of adopting the BPA on SAT: the iteration of propagation is independent to the size of the SAT problem. Also, as mentioned above, each computation of r and q only depends on the previous iteration result of q and r, which can be implemented in parallel easily with a buffer storing its previous value. Moreover, the sparse literal-clause matrix release the competing of the shared memory issue, which is applicable to implement on both FPGA, GPU, or even ASIC devices to reduce the synthesis timing overhead.

## 5. RELATED WORK

Application-specific architecture is the mainstream in hardware accelerating SAT problem to avoid the time-consuming FPGA synthesis time. [10][11] both implemented Boolean Constraint Propagation computation in parallel. For the storage on the literal-clause instance, the former stored the data using embedded DRAM, while the latter relied on the modern FPGA's Block RAM (BRAM). Both of them relied on Input/Output Queue to provide parallel value assignment. The intensive BCP computation was done in parallel by partitioning clause set into several subset. The synchronization scheme is also proposed to do the conflict detection in serial. In this type of architecture, the problem size can be very large. For latter one, since the size of BRAM is the only constraint of this hardware implementation, the capacity can be extended to 64K variables and 176K clauses.

Skliarova *et al.* [7] implemented the application-specific architecture by storing the whole literal-clause matrix into the FPGA block. The inference, conflict-detection, and backtracking are implemented based on the orthogonal matrix multiplication, which all the computations are done in hardware implementation, while the backstracking process is supported by the stack memory handled by software.

## 6. CONCLUSION

In this paper, a belief propagation algorithm for CNF-based SAT problems (BPA-SAT) has been presented. The BPA-SAT is based on the belief propagation algorithm (BPA) that is used in the decoding of a class of error-correcting codes, low-density parity-check (LDPC) codes. Since BPA can be implemented with hardware, the BPA-SAT solver can be accelerated by hardware. Furthermore, the experimental results show that the time complexity of the BPA-SAT does not increase with the size of SAT problems. Therefore, the speed solver does not degrade dramatically as the size of the CNF increases. The main issue, completeness, of this algorithm is also investigated. The completeness decreases significantly as the size of the CNF increases. The experimental results and our estimation show that for the instants that the BPA-SAT can solve, BPA-SAT can achieve 30× to 44k× speedup compared to a state-of-art SAT solver, MiniSat.